\documentclass[letterpaper, 10 pt,journal, twoside]{IEEEtran}  

\IEEEoverridecommandlockouts                              

\usepackage{graphics} 
\usepackage{epsfig} 
\usepackage{times} 
\usepackage{amsmath} 
\usepackage{amssymb}  
\usepackage{amsfonts}
\usepackage[colorlinks,linkcolor=blue,anchorcolor=blue,citecolor=blue]{hyperref}
\usepackage[table]{xcolor}
\usepackage{booktabs}
\usepackage{multirow}
\usepackage{graphicx}
\usepackage{subfigure}
\usepackage{cite}
\usepackage{float}
\usepackage{makecell}
\usepackage{algorithm}
\usepackage{colortbl}
\usepackage{xcolor}
\definecolor{aggro_yellow}{RGB}{201, 141, 0} 
\definecolor{cons_blue}{RGB}{0, 120, 186}
\definecolor{norm_red}{RGB}{240, 101, 97}
\usepackage[noend]{algorithmic}
\usepackage{etoolbox}
\usepackage[flushleft]{threeparttable}
\hypersetup{
    colorlinks=true,
    citecolor=blue    
}

\makeatletter
\patchcmd{\@makecaption}
  {\scshape}
  {}
  {}
  {}
\makeatletter
\patchcmd{\@makecaption}
  {\\}
  {.\ }
  {}
  {}
\makeatother
\def\tablename{Table}

\graphicspath{{img/}}

\title{D$^3$-MoE: Dual Disentangled Diffusion Mixture-of-Experts  for Style-Controllable End-to-End Autonomous Driving}
 \vspace{-0.5cm}
\author{Renju Feng$^{\dagger}$, Rukang Wang$^{\dagger}$,  Ning Xi, Jianguo Yu, Liping Lu, Pan Zhou, and Duanfeng Chu
\thanks{This work is supported in part by the National Natural Science Foundation of China (52472438), the Key R\&D Program of Hubei Province (2024BAB033, 2025BAB081), the Science and Technology Project of the Department of Transport of Hubei Province (2025-69-3-5), the Wuhan Joint R\&D Project Under the National Transportation Power Construction Pilot Program (2025-2-7), and the Wuhan Municipal key R\&D Program (20250512030403).
(Renju Feng and Rukang Wang contributed equally to this work.) ~(\textit{Corresponding author: Duanfeng Chu.})}
\thanks{Renju Feng, Rukang Wang, Ning Xi, and Duanfeng Chu are with the Intelligent Transportation Systems Research Center, Wuhan University of Technology, Wuhan,  Hubei, China. (e-mail: \href{mailto:fengrenju@whut.edu.cn}{\texttt{\small fengrenju@whut.edu.cn}}, \hspace{-0.5em}\href{mailto:wangrk@whut.edu.cn}{\texttt{\small wangrk@whut.edu.cn}}, \hspace{0.25em}\href{mailto:xining095@whut.edu.cn}{\texttt{\small xining095@whut.edu.cn}},\hspace{0.25em}
 \href{mailto:chudf@whut.edu.cn}{\texttt{\small chudf@whut.edu.cn}})} 
\thanks{Jianguo Yu is with the School of Mechanical and Electronic Engineering, Wuhan University of Technology, Wuhan,  Hubei, China. (e-mail: \href{mailto:yujg@whut.edu.cn}{\texttt{\small yujg@whut.edu.cn}})}
  \thanks{Liping Lu is with the School of Computer Science and Artificial Intelligence, Wuhan University of Technology, Wuhan, Hubei, China. (e-mail: \href{mailto:luliping@whut.edu.cn}{\texttt{\small luliping@whut.edu.cn}})}
   \thanks{Pan Zhou is with Hubei Key Laboratory of Distributed System Security,
School of Cyber Science and Engineering, Huazhong University of Science
and Technology, Wuhan, Hubei, China. (e-mail: \href{mailto: panzhou@hust.edu.cn}{\texttt{\small  panzhou@hust.edu.cn}})}
   }

\begin{document}

\maketitle
\pagestyle{empty} 
\thispagestyle{empty} 

\begin{abstract}
Traditional end-to-end autonomous driving frameworks frequently suffer from the ``style-averaging'' dilemma when trained on high-variance human demonstrations, yielding homogenized, style-uncontrollable, and even kinematically unsafe policies.
To overcome this limitation, we present $\textbf{D}^3\textbf{-MoE}$ (Dual Disentangled Diffusion Mixture-of-Experts), which disentangles trajectory modeling along two complementary axes. On the behavioral axis, generation is decoupled from selection: a style-conditioned diffusion process synthesizes multi-style candidate trajectories in parallel within a single scene, allowing a downstream module to select the optimal trajectory based on user preference or an evaluation score. 
On the physical axis, decoupled longitudinal and lateral routers activate their respective experts during inference time, trained without manual labels using self-supervised targets from orthogonal ground-truth kinematics.
These activated experts, architected as Diffusion Transformers (DiT) and equipped with style-conditioned AdaLN and asymmetric lateral-fusion cross-attention, independently predict their corresponding physical state before being reassembled into a unified, kinematically coherent trajectory.
Extensive evaluations on the challenging NAVSIM benchmark demonstrate that $\textbf{D}^3\textbf{-MoE}$ achieves state-of-the-art planning performance, reaching 88.2 PDMS and 84.3 EPDMS by default. Moreover, our \textit{Best-of-Three} ensemble strategy effectively broadens the multi-modal solution space, raising performance to 91.3 PDMS and 87.5 EPDMS. Both quantitative and qualitative analyses jointly confirm the framework's advantages in planning quality and style controllability.
\end{abstract}
 \vspace{-0.5cm}

\section{Introduction}
\IEEEPARstart{E}{nd}-to-end autonomous driving has rapidly emerged as the prevailing paradigm for modern self-driving systems, overcoming the cascading errors and constrained intermediate representations inherent in traditional modular pipelines \cite{li2024bevformer,xin2025multi,pei2026safe}. By directly mapping multi-sensor observations to planning outputs under a unified optimization objective, these models natively capture complex interaction cues and decision dependencies in dynamic traffic environments \cite{hu2023uniad,chitta2022transfuser,feng2025artemis}.

While imitation learning underpins modern end-to-end planning \cite{chen2024end}, traditional regression-based approaches struggle to capture the multi-modal distribution of human driving behavior \cite{wu2026unified,zheng2026unleashing}. The root cause is a structural mismatch between real-world variance and dataset supervision. While perceptually similar scenarios across a dataset often justify highly divergent maneuvers, such as aggressive passing versus conservative yielding, offline datasets record only a single demonstration for any given situational context. Mapping these similar inputs to conflicting expert actions creates a pathological conflict at the objective level. Because standard regression losses are Bayes-optimal at the conditional mean of the targets, the network is provably driven toward the statistical \emph{average} of these valid modes\cite{florence2022implicit} . This average often lies outside the set of physically reasonable maneuvers, producing a dangerous, half-committed motion between yielding and overtaking. The resulting ``style averaging'' is therefore not an engineering oversight but a fundamental mathematical property of the loss function, degrading the planner into homogenized policies that erase legitimate behavioral diversity\cite{chi2025diffusion}.

To escape the style-averaging trap, recent approaches leverage diffusion models \cite{dang2026drivefine,liao2024diffusiondrive,wang2026feaxdrive, zou2025diffusiondrivev2,liu2025guideflow}. Rather than regressing a single point estimate, these models utilize an iterative denoising paradigm to approximate the full conditional distribution. By sampling directly from this distribution, diffusion models inherently preserve distinct behavioral modes (e.g., passing versus yielding), generating dynamically feasible and highly diverse trajectories \cite{gui2025trajdiff,song2025diver}. Yet, eliminating style averaging exposes two new critical bottlenecks. Architecturally, compressing the unbounded spectrum of open-world scenarios into a monolithic diffusion network forces conflicting maneuvers to compete for shared weights, empirically inducing severe gradient interference that systematically degrades rare, safety-critical behaviors \cite{feng2025artemis,yuan2025pie}.  
Behaviorally, standard diffusion models generate trajectories by unconditional stochastic sampling, exposing no explicit interface to specify which behavioral mode is produced or to deterministically select the one to execute. This lack of controllability is unacceptable in safety-critical driving, where deployment demands explicit, style-controllable planning to satisfy varying passenger comfort profiles and context-specific safety constraints.

\begin{figure}
    \centering
    \includegraphics[width=\columnwidth]{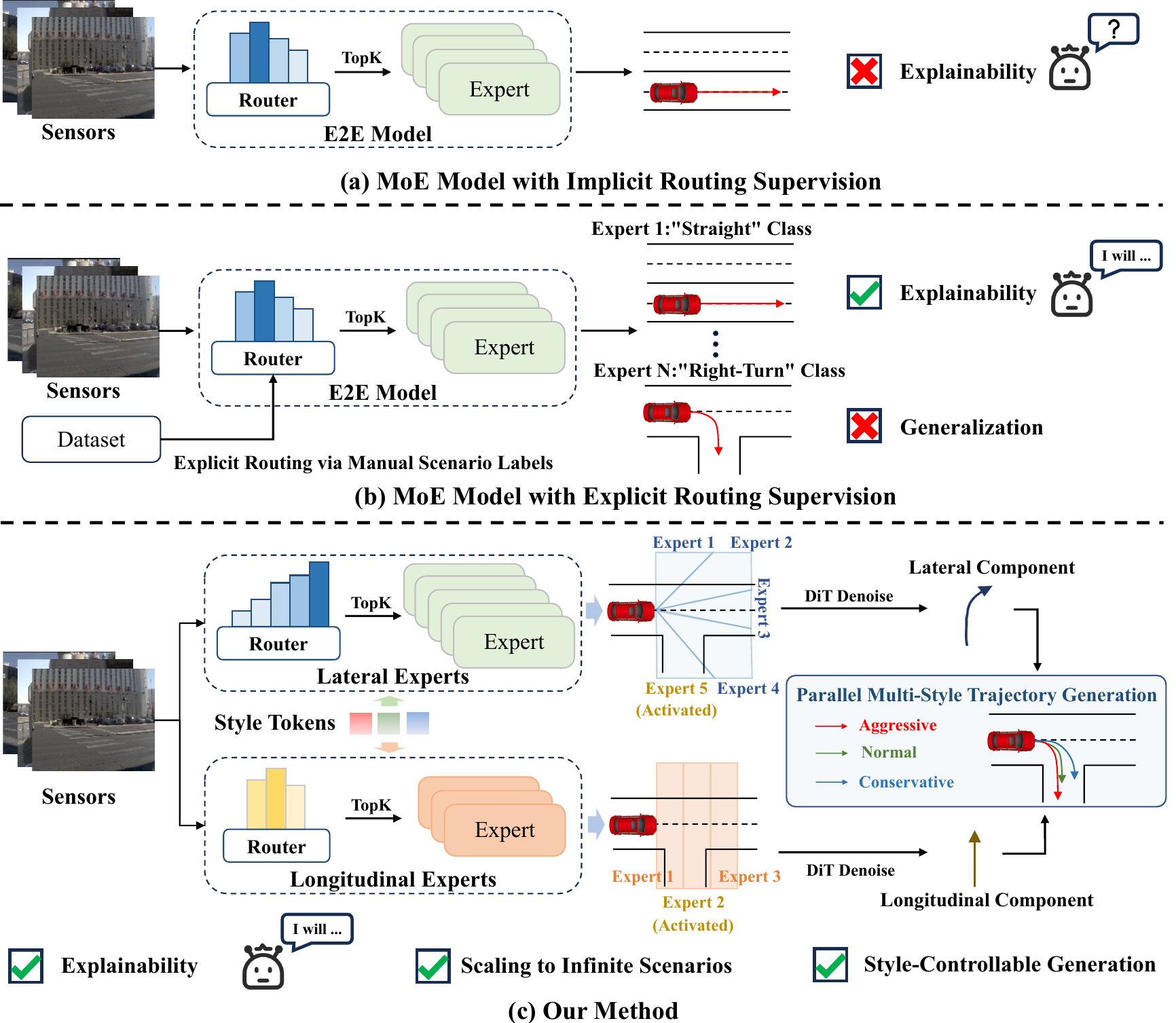}
    \vspace{-0.5cm}
    \caption{\textbf{Comparison of MoE architectures.} 
    (a) MoE models with implicit routing supervision \cite{feng2025artemis,sun2025generalizing,xing2025trajmoe,yuan2025pie} inherently lack physical interpretability. (b) MoE models relying on explicit routing via manual scenario labels \cite{wan2025geminus,xu2025mose} struggle to generalize across open-world environments. (c) Our proposed \textbf{D}$^3$\textbf{-MoE} framework compresses unbounded scenarios into combinations of finite, orthogonal physical primitives by explicitly decoupling lateral and longitudinal generation. Supervised purely by ground-truth kinematics rather than manual labels, independent routers activate specialized Diffusion Transformer (DiT) experts during inference. Equipped with style-conditioned AdaLN and asymmetric lateral-fusion cross-attention, these experts independently denoise their assigned physical states, which are subsequently reassembled to enable multi-style, kinematically coherent trajectory generation.}
    \vspace{-0.5cm}
    \label{fig:introduction}
\end{figure}

Consequently, Mixture of Experts (MoE) systems have emerged as a primary strategy to alleviate the representational limits of monolithic networks \cite{jiang2026expertad,feng2025artemis,xing2025trajmoe,wan2025geminus,xu2025mose}. These architectures aim to synergistically enhance scenario adaptability via specialized sub-networks. However, constructing MoE frameworks that maintain strong physical interpretability and support style control remains fundamentally challenging. The efficacy of an MoE architecture heavily depends on its routing mechanism. Relying on dataset-provided scene labels \cite{wan2025geminus} or manual annotations \cite{xu2025mose} yields rigid, discrete assignments that fail to generalize across the continuous, inexhaustible spectrum of open-world long-tail events. Conversely, relying purely on implicit routing mechanisms (often regularized only by load-balancing constraints) \cite{feng2025artemis,sun2025generalizing,xing2025trajmoe,yuan2025pie} results in entangled expert specializations that lack physical meaning, severely degrading system controllability.

Motivated by these compounding limitations, we propose \textbf{D}$^3$\textbf{-MoE} (Dual-Disentangled Diffusion Mixture-of-Experts ), an end-to-end planning framework that disentangles trajectory modeling along two complementary axes, as illustrated in Fig.~\ref{fig:introduction}. On the \emph{behavioral} axis, trajectory generation is explicitly decoupled from selection: a style-conditioned diffusion process synthesizes a diverse pool of multi-style candidate trajectories in parallel, allowing a downstream module to select the optimal path based on user preference or learned scores. This fundamentally escapes the style-averaging trap of unimodal regression. 
On the \emph{physical} axis, decoupled longitudinal and lateral routers activate their respective experts during inference, having been trained without manual labels using self-supervised targets from orthogonal ground-truth kinematics. These activated experts, architected as Diffusion Transformers (DiT) and equipped with style-conditioned AdaLN and asymmetric lateral-fusion cross-attention, independently predict their corresponding physical states before being reassembled into a unified, kinematically coherent trajectory. Ultimately, \textbf{D}$^3$\textbf{-Mo3} successfully synergizes MoE specialization with the multi-modal expressivity of generative diffusion models, providing explicit, fine-grained control across the behavioral spectrum from aggressive to conservative, enabling  physically feasible, user-centric autonomous driving.

Our contributions can be summarized as follows:

\textbf{(1)} We propose \textbf{D}$^3$\textbf{-MoE}, an architecture that explicitly decouples trajectory generation from selection. By maintaining a diverse pool of lateral and longitudinal experts, it overcomes the conventional ``style-averaging'' dilemma and enables style-controllable trajectory planning.

\textbf{(2)} We design a  diffusion Mixture-of-Experts framework with two key components. First, decoupled longitudinal and lateral routers activate their respective experts during inference, having been trained annotation-free via self-supervised targets derived from orthogonal ground-truth kinematics. Second, these activated experts, architected as Diffusion Transformers (DiT) with style-conditioned AdaLN and asymmetric lateral-fusion cross-attention, independently predict their physical states before being reassembled into a unified, kinematically coherent trajectory.

\textbf{(3)} Extensive evaluations on the NAVSIM \texttt{navtest} benchmark validate our approach. Under the default setting, \textbf{D}$^3$\textbf{-MoE} achieves a competitive 88.2 PDMS and 84.3 EPDMS. Crucially, utilizing our \textit{Best-of-Three} ensemble strategy further raises the performance  to 91.3 PDMS and 87.5 EPDMS, demonstrating comprehensive superiority in planning quality and multi-modal trajectory exploration.

\vspace{-0.2cm}
\section{Method}
\begin{figure*}
    \centering
    \includegraphics[width=1\textwidth]{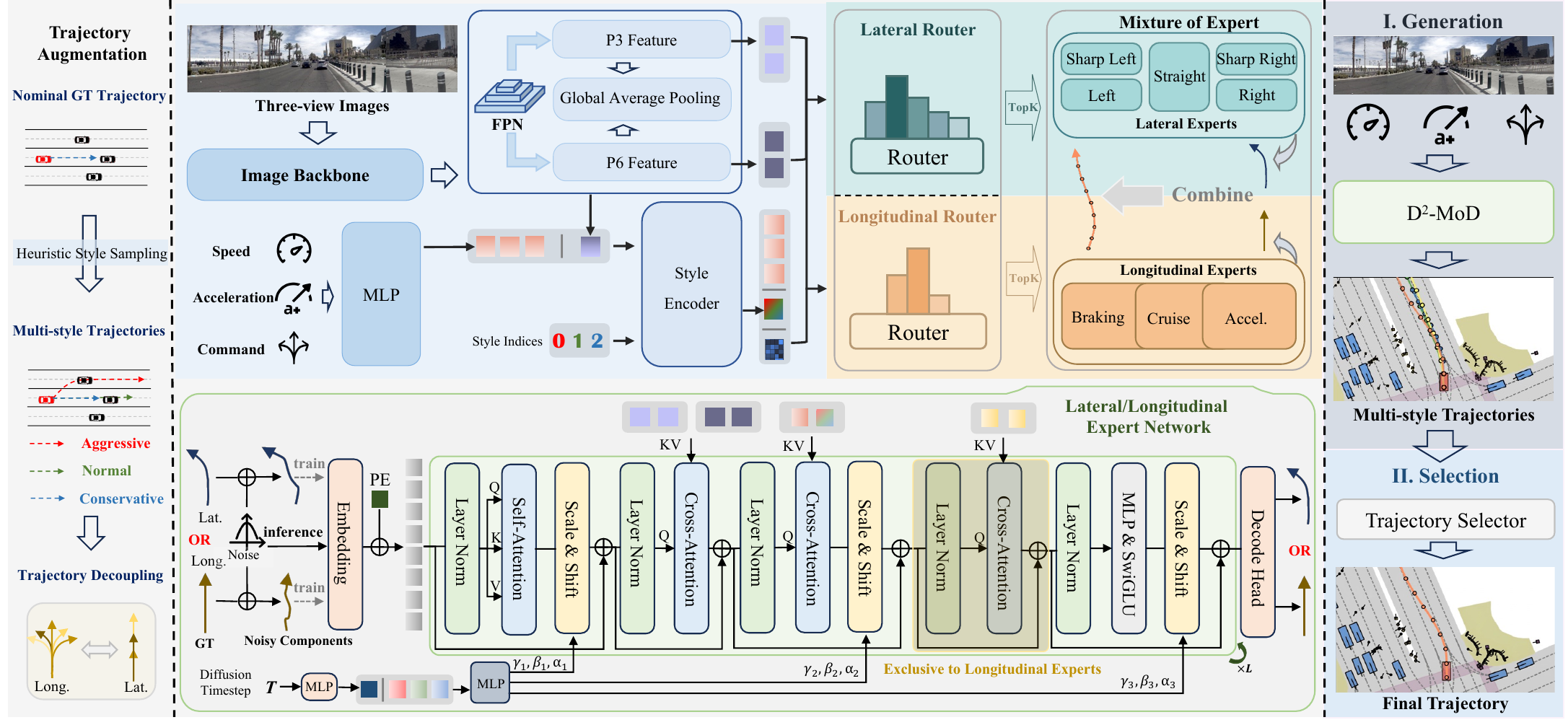}
    \vspace{-0.5cm}
    \caption{\textbf{Methodological pipeline of D$^3$-MoE.} \emph{Left}: Offline multi-style expert reference trajectory augmentation. \emph{Middle}: The core denoising architecture with dynamic routing and independent lateral/longitudinal DiT experts. \emph{Right}: The behavioral decoupling paradigm for parallel multi-style trajectory generation and their subsequent downstream selection,  which features a \textbf{user-centric} mode driven by language instructions or style indices, and a \textbf{score-centric} mode that generates parallel multi-style candidates for downstream evaluation and optimal selection.}
    \vspace{-0.3cm}
    \label{fig:framework}
\end{figure*}

To resolve the dual bottlenecks of representational capacity and style controllability in end-to-end autonomous driving, we propose $\textbf{D}^3\textbf{-MoE}$ (Dual-Disentangled Diffusion Mixture-of-Experts). As illustrated in Fig.~\ref{fig:framework}, 
the framework operates through a strictly decoupled pipeline. The generative target is orthogonally decomposed into lateral and longitudinal dimensions, where dedicated routers dynamically activate independent experts for style-guided denoising to synthesize parallel multi-style candidates.
Generation is further decoupled from selection via two downstream paradigms: a user-centric mode driven by language or explicit style indices, and a score-centric mode utilizing an independent evaluator to rank and select the optimal trajectory.

\vspace{-0.4cm}
\subsection{Problem Formulation}
We formulate the end-to-end planning of the ego vehicle's trajectory as a conditional diffusion generation problem. 
Anchored at the current ego-state in the ego-centric frame, the future trajectory $\tau = \{(x_k, y_k, \theta_k)\}_{k=1}^K$ is orthogonally decoupled into a longitudinal displacement sequence $\tau_{lon} = \{\Delta x_k\}_{k=1}^K$ and a lateral sequence $\tau_{lat} = \{(\Delta y_k, \Delta\theta_k)\}_{k=1}^K$ representing lateral and heading changes.

To capture the multi-modal nature of driving behavior, both the lateral and longitudinal domains are equipped with a dedicated ensemble of independent experts: five lateral experts (\textit{sharp left, left, straight, right, sharp right}) and three longitudinal experts (\textit{braking, cruising, accelerating}). Two decoupled routers then dynamically activate the optimal expert  to generate the corresponding trajectory components.

In each forward pass, the activated lateral and longitudinal experts operate as independent denoisers to reconstruct their respective spatial dimensions from the noisy input $\tau_t$ at timestep $t$ under condition $C$. Deviating from standard $\epsilon$-prediction, we implement an $x_0$-prediction parameterization, training each expert to directly regress the clean ground-truth trajectory components for its assigned axis, which collectively form the final trajectory $\hat{\tau}_0$.

\vspace{-0.4cm}
\subsection{Dynamic Routing}
Driven by a dynamic routing network, $\textbf{D}^3\textbf{-MoE}$ spatially decouples expert allocation. By leveraging the kinematics of ground-truth trajectories for self-supervised training, the framework completely eliminates the need for manual labels while ensuring highly interpretable expert matching across diverse scenarios. 

\textbf{Routing Network Architecture.}
The routing process begins with multi-scale scene perception. A VoVNetV2-99-eSE backbone extracts a feature pyramid from the input image $\mathbf{I} \in \mathbb{R}^{3 \times H \times W}$, which is fused via a Feature Pyramid Network (FPN) \cite{lin2017feature} and projected to a unified channel dimension $D_\text{context}$. With spatial topology preserved via convolutional positional encodings, the feature grids are flattened into a contiguous visual sequence $\mathcal{F}_\text{vision} \in \mathbb{R}^{N_\text{vis} \times D_\text{context}}$. Concurrently, global average pooling over $\mathcal{F}_\text{vision}$ yields a compact macroscopic scene representation $\mathbf{F}_\text{global} \in \mathbb{R}^{D_\text{context}}$.

Alongside the visual context, the current ego-state is MLP-encoded into a dense vector $\mathbf{e} \in \mathbb{R}^{D_\text{context}}$. To capture distinct driving behaviors, a style encoder concatenates $\mathbf{F}_\text{global}$ and $\mathbf{e}$ to generate the $i$-th style embedding $\mathbf{s}_i$. A state-fusion Transformer then constructs a unified query sequence $\mathbf{Q}_{\text{state}, i}$ by processing the concatenated tokens $[\mathbf{e}, \mathbf{s}_i, \text{cls}]$, where $\text{cls}$ acts as a  learnable class token.

To ground this intent spatially, a cross-attention module extracts specific multi-scale visual cues from selected levels of the feature pyramid ($j \in \{3, 6\}$):
 \vspace{-0.1cm}
\begin{equation}
\mathbf{z}_{i,j} = \text{CrossAttn}(\mathbf{Q}_{\text{state}, i}, \mathcal{F}_\text{vision}^{(j)})
\end{equation}
The aggregated spatial-semantic context $\mathbf{Z}_i = [\mathbf{z}_{i,3}, \mathbf{z}_{i,6}]$ is subsequently fed into the decoupled decision heads, yielding independent routing logits for the lateral and longitudinal axes:
 \vspace{-0.2cm}
\begin{equation}
\mathbf{O}_{\text{lat}, i} = \Phi_\text{lat}(\mathbf{Z}_i), \qquad \mathbf{O}_{\text{lon}, i} = \Phi_\text{lon}(\mathbf{Z}_i)
\end{equation}
Finally, Softmax normalization converts these logits into continuous gating probabilities, driving a Top-$k$ selection mechanism to discretely activate the optimal independent experts.

\textbf{Self-Supervised Training via Trajectory Kinematics.}
To train the routing network without costly manual annotations, we extract self-supervised soft labels directly from ground-truth trajectories $\tau_{GT}$. Specifically, we define a four-dimensional kinematic feature vector $\mathcal{P}$, decoupled into two lateral descriptors (net heading change $\Delta\theta$, signed
maximum lateral displacement $d_{lat}$) and two longitudinal descriptors
(average speed $\bar{v}$, average acceleration $\bar{a}$).

Given standardized features $\mathcal{P}'$, independent $K$-means clustering along each axis yields cluster centers $\mathbf{C}_k$ and standard deviations $\boldsymbol{\sigma}_k$. The soft routing target $\hat{y}_k$ is computed via a Mahalanobis-based softmax:
\begin{equation}
\hat{y}_k =\frac{\exp\!\left(-\tfrac{1}{2\kappa}\sum_{d}\dfrac{(\mathcal{P}'_d - C_{k,d})^2}{\sigma_{k,d}^2}\right)}     {\sum_{j}\exp\!\left(-\tfrac{1}{2\kappa}\sum_{d}\dfrac{(\mathcal{P}'_d - C_{j,d})^2}{\sigma_{j,d}^2}\right)},
\end{equation}
where $\mathcal{P}'_d$, $C_{k,d}$, and $\sigma_{k,d}$ denote the standardized feature, cluster center, and standard deviation along dimension $d$. Temperature $\kappa$ controls assignment smoothness, independently yielding the decoupled targets $\hat{\mathbf{y}}_\text{lat}$ and $\hat{\mathbf{y}}_\text{lon}$."

The routing network is end-to-end optimized by minimizing the KL divergence between its predicted gating probabilities $\mathbf{G}_a$ and these physics-derived labels:
\begin{equation}
\mathcal{L}_\text{route} = \sum_{a \in \{\text{lat},\, \text{lon}\}} D_\text{KL}\!\left(\hat{\mathbf{y}}_a \,\big\|\, \mathbf{G}_a\right)
\end{equation}

 \vspace{-0.8cm}
\subsection{Style-Aware MoE Diffusion}
Once the routers selectively activate the appropriate experts, these independent networks transform Gaussian noise into decoupled trajectory components, which are subsequently combined into a physically feasible, style-aligned trajectory. 

\vspace{1ex}
\textbf{Denoising Experts.}
At timestep $t$, the independent forward processes yield decoupled noisy inputs $\tau_{\text{lat}, t}$ and $\tau_{\text{lon}, t}$. Each sequence is embedded, added to a positional encoding $\mathbf{PE}$, and processed by $L$ stacked DiT blocks. Within the $l$-th block, hidden features undergo Self-Attention, Multi-Scale Visual Cross-Attention, Unified Condition Cross-Attention, and a SwiGLU-activated MLP. All sub-modules utilize standard LayerNorm and residual connections. The key and value for $\mathcal{C}_\text{uni}$ originate from $\mathbf{C}_\text{unified} = [\,\tilde{\mathbf{e}};\; \tilde{\mathbf{s}}_i\,] \in \mathbb{R}^{2 \times D_\text{context}}$, constructed by concatenating the independent 2-layer MLP projections of ego-state $\mathbf{e}$ and style feature $\mathbf{s}_i$.

To ensure kinematic coherence, the longitudinal expert features an exclusive asymmetric lateral-fusion cross-attention before the MLP. Attending strictly to the lateral expert's corresponding hidden states $\mathbf{H}_\text{lat}^{(l)}$, it conditions longitudinal acceleration directly on the lateral path geometry. Finally, decoder heads map the terminal features into their respective sequences: $\tau_\text{lon} = \{\Delta x_k\}_{k=1}^K$ or $\tau_\text{lat} = \{(\Delta y_k, \Delta\theta_k)\}_{k=1}^K$.

\vspace{1ex}
\textbf{Adaptive Injection of Style Conditions.} 
The block-wise Adaptive Layer Normalization (AdaLN) is driven by a global style vector $\mathbf{s}_\text{global} \in \mathbb{R}^{D_\text{context}}$, derived from the style feature $\mathbf{s}_i$ via an MLP. Concatenating $\mathbf{s}_\text{global}$ with the sinusoidal timestep embedding $\mathbf{t}_\text{emb}$ enables a two-layer SiLU-activated MLP to yield nine modulation parameters per block.
These parameters comprise the scale $\gamma$, shift $\beta$, and residual gate $\alpha$ for the three sub-modules, thereby jointly conditioning the feature flow on time and driving style.

\textbf{Expert Component Loss ($\mathcal{L}_\text{expert}$).} 
To preclude semantic entanglement during denoising, we apply layer-wise deep supervision. The intermediate trajectory predictions from all $L$ blocks are directly constrained against their respective clean ground truths via an $L_1$ penalty:
 \vspace{-0.3cm}
\begin{equation}
\mathcal{L}_\text{expert} = \sum_{l=1}^{L} w_l \left( \| \hat{\tau}_\text{lat}^{(l)} - \tau_\text{lat} \|_1 + \| \hat{\tau}_\text{lon}^{(l)} - \tau_\text{lon} \|_1 \right)
\end{equation}
where $w_l$ represents the predefined weight for the $l$-th layer.

\vspace{-0.4cm}
\subsection{Unified Diffusion Training Objective}
The predicted longitudinal $\tau_\text{lon} = \{\Delta x_k\}_{k=1}^K$ and lateral $\tau_\text{lat} = \{(\Delta y_k, \Delta\theta_k)\}_{k=1}^K$ sequences are reassembled into the complete trajectory $\hat{\tau}$. Following the $\mathbf{x}_0$-prediction formulation, this output is supervised against the ground truth $\tau_\text{GT}$ using a hybrid $L_1$-$L_2$ reconstruction loss:
\begin{equation}
\mathcal{L}_\text{traj} = \lambda_\text{L1} \| \hat{\tau} - \tau_\text{GT} \|_1 + \lambda_\text{L2} \| \hat{\tau} - \tau_\text{GT} \|_2^2
\end{equation}

During end-to-end training, the model minimizes a weighted combination of the three loss terms. Since the routing objective $\mathcal{L}_\text{route}$ is supervised solely by the ground-truth kinematics and operates independently of the diffusion process, the expectation over the diffusion timestep $t \sim \mathcal{U}(1, T)$ and the sampled Gaussian noise $\epsilon \sim \mathcal{N}(\mathbf{0}, \mathbf{I})$ applies exclusively to the diffusion-based components:
\begin{equation}
\mathcal{L}_\text{total} =\! \mathbb{E}_{\tau_\text{GT}} \!\Big[ \lambda_\text{route} \mathcal{L}_\text{route} +\! \mathbb{E}_{t, \epsilon} \!\big[ \lambda_\text{traj} \mathcal{L}_\text{traj} + \lambda_\text{expert} \mathcal{L}_\text{expert} \big] \Big]
\end{equation}

\vspace{-0.4cm}
\section{Experiments}

\subsection{Dataset and Stylized Reference Trajectory Augmentation}

\textbf{Dataset Setup.} We train and evaluate our model on the NAVSIM~\cite{dauner2024navsim} dataset. Each sample provides multi-modal inputs, including multi-view camera images ($8$ cameras), fused LiDAR point clouds ($5$ sensors), map annotations, and 3D bounding boxes. 
Given 4 historical frames spanning the preceding 2 seconds, the model
predicts an 8-waypoint trajectory over a 4-second planning horizon.

\begin{figure}[ht]
\centering
\includegraphics[width=\columnwidth]{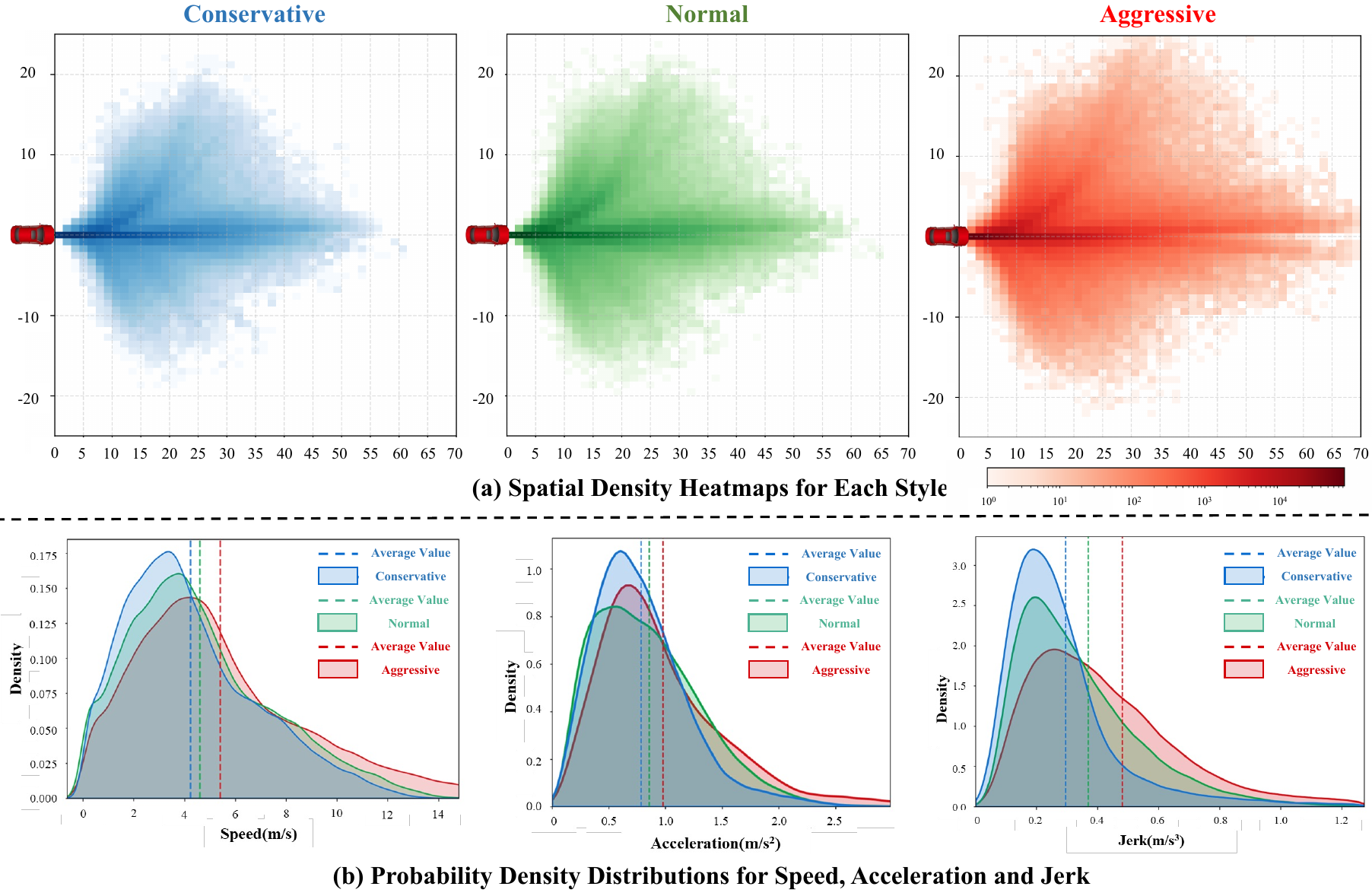}
\vspace{-0.5cm}
\caption{\textbf{Distributions of the stylized reference trajectories.} (a) Spatial density heatmaps of the different stylic trajectories. (b) Probability density distributions for speed, acceleration, and jerk. Aggressive trajectories shift toward higher speed, acceleration, and jerk,
while Conservative ones concentrate at lower magnitudes, confirming
kinematically distinct styles.}
\vspace{-0.3cm}
\label{fig:command}
\end{figure}

\vspace{1ex}
\textbf{Multi-Style Trajectory Augmentation.}
For explicit multi-modal style supervision, we expand each NAVSIM sample into a tri-modal trajectory pool. The ground-truth (GT) trajectory serves as the \textit{Normal} baseline, while \textit{Aggressive} and \textit{Conservative} counterparts are synthesized offline. For each style, we generate candidates via heuristic mutations, filter them for kinematic feasibility, and evaluate them using a closed-loop scorer. The top-scoring candidate is paired with the GT to finalize the tri-modal expert set (see Fig.~\ref{fig:command}).

\textit{\textbf{1) Aggressive Synthesis:}} We amplify the GT velocity profile by $1.05$--$1.25\times$ and spatially re-extend it along the ego-lane centerline. To prevent unrealistic acceleration, speeds are dynamically clipped by local curvature and posted limits, while initial states are smoothly blended for kinematic continuity. 

\textit{\textbf{2) Conservative Synthesis:}} We employ an escalating safety cascade: (i) reducing GT speed and widening the car-following margin; (ii) if suboptimal, enforcing a strict \emph{deceleration and lane-centering} fallback; and (iii) ultimately truncating the planning horizon to emulate maximum caution.

\vspace{-0.3cm}
\subsection{Evaluation Metrics}
We adopt the Predictive Driving Model Score (PDMS)~\cite{dauner2024navsim} and its extended variant (EPDMS)~\cite{li2025hydra++} as our primary evaluation metrics. 
Specifically, the base PDMS assesses five critical driving dimensions: No Collisions (NC), Time-to-Collision (TTC), Driveable Area Compliance (DAC), Comfort (C), and Ego Progress (EP).

To further evaluate the model's adherence to fine-grained traffic regulations, we report the EPDMS (Extended PDMS)~\cite{li2025hydra++}. Building upon the base metric, EPDMS incorporates three additional soft constraints: Lane Keeping (LK), Historical Comfort (HC), and Extended Comfort (EC). Crucially, it also applies strict multiplicative penalties for severe safety infractions, specifically targeting Driving Direction Compliance (DDC) and Traffic Light Compliance (TLC).

\vspace{-0.4cm}
\subsection{Implementation Details}
We instantiate our decoupled MoE architecture with $E_{\text{lat}}=5$ lateral and $E_{\text{lon}}=3$ longitudinal experts, activating a single top-1 expert per branch ($K_{\text{lat}}=K_{\text{lon}}=1$) during both training and inference. The model is trained on the \texttt{navtrain} split using a single NVIDIA RTX 4090 GPU. All detailed hyperparameters are comprehensively summarized in Table \ref{tab:implementation_details}.
\vspace{-0.6cm}
\begin{table}[!htb]
\centering
\small
\caption{Hyperparameter Configurations.}
\label{tab:implementation_details}
\begin{tabular}{llc}
\toprule
\textbf{Parameter} & \textbf{Symbol} & \textbf{Value} \\
\midrule
Lateral Experts & $E_{\text{lat}}$ & 5 \\
Longitudinal Experts & $E_{\text{lon}}$ & 3 \\
Active Experts (Top-1) & $K_{\text{lat}}, K_{\text{lon}}$ & 1 \\
Training Epochs & - & 100 \\
Batch Size & - & 32 \\
Initial Learning Rate & - & $1 \times 10^{-4}$ \\
Weight Decay & - & $1 \times 10^{-4}$ \\
Diffusion Denoising Steps & - & 2 \\
Layer Weights & $w_1, w_2, w_3$ & 0.3, 0.65, 1.0 \\
L1 / L2 Loss & $\lambda_{L1}$ / $\lambda_{L2}$ & 10 / 5 \\
Routing Loss & $\lambda_{\text{route}}$ & 10 \\
Expert Loss & $\lambda_{\text{expert}}$ & 20 \\
Trajectory Loss & $\lambda_{\text{traj}}$ & 1 \\
\bottomrule
\end{tabular}
\end{table}

\vspace{-0.4cm}

\section{Results}

\subsection{Quantitative Results}
\textbf{PDMS Comparison.} We evaluate \textbf{D}$^3$\textbf{-MoE} against representative methods on the Navtest benchmark (Tab.~\ref{tab:navsim_v1}). The \textit{Normal} style achieves a state-of-the-art PDMS of 88.2, dominating key metrics like NC, DAC, and EP. The \textit{Conservative} (85.5 PDMS) and \textit{Aggressive} (80.9 PDMS) branches exhibit expected design trade-offs: the former sacrifices efficiency (EP 79.6) for safety but occasionally suffers progress collapse in long-tail scenarios, while the latter incurs heavy safety penalties (NC/DAC/TTC drops) due to radical maneuvers. Notably, an \textit{Best-of-Three} selection boosts PDMS to 91.3, demonstrating the vast upper-bound potential of our tri-modal diverse proposals.
\begin{table}[ht]
  \centering
  \caption{\textbf{Base Metrics on Navtest.}}
  \label{tab:navsim_v1}
  \footnotesize 
  \setlength{\tabcolsep}{2pt}
  \begin{tabular}{lc|ccccc>{\columncolor{gray!20}}c}
    \toprule
    \textbf{Method}&\textbf{Input} & \textbf{NC$\uparrow$} & \textbf{DAC$\uparrow$} & \textbf{EP$\uparrow$} & \textbf{TTC$\uparrow$} & \textbf{C$\uparrow$} & \textbf{PDMS$\uparrow$} \\
    \midrule
    VADv2~\cite{chen2024vadv2} &C\&L& 97.2 & 89.1 & 76.0 & 91.6 & \textbf{100} & 80.9 \\
    UniAD~\cite{hu2023uniad} &C& 97.8 & 91.9 & 78.8 & 92.9 & \textbf{100} & 83.4 \\
      Transfuser~\cite{chitta2022transfuser}&C\&L & 97.7 & 92.8 & 79.2 & 92.8 & \textbf{100} & 84.0 \\
    PARA-Drive~\cite{weng2024para} &C& 97.9 & 92.4 & 79.3 & 93.0 & 99.8 & 84.0 \\
    DriveX\cite{shi2025drivex} &C& 97.5 & 94.0 & 79.7 & 93.0 & \textbf{100} & 84.5 \\
      LAW\cite{li2024enhancing} &C  & 96.4 & 95.4 & 81.7 & 88.7 & \underline{99.9} & 84.6 \\
      FSDrive\cite{zeng2025futuresightdrive} &C  & \underline{98.2} & 93.8 & 80.1 & 93.3 & \underline{99.9} & 85.1 \\
      NoRD~\cite{rawal2026nord} & C\&L & 97.6 & 94.9 & 79.3 & 93.5 & \textbf{100} & 85.6 \\
    Epona~\cite{zhang2025epona} &C & 97.9 & 95.1 & 80.4 & 93.8 & \underline{99.9} & 86.2 \\
    DistillDrive~\cite{yu2025distilldrive} &C\&L & 98.1 & 94.6 & 81.0 & 93.6 & \textbf{100} & 86.2 \\
    ARTEMIS~\cite{feng2025artemis} &C\&L & \textbf{98.3} & 95.1 & 81.4 & 94.2 & \textbf{100} & 86.9 \\
    PRIX~\cite{wozniak2026prix} & C & 98.1 & \underline{96.3} & \underline{82.3} & 94.1 & \textbf{100} & 87.8 \\
    DiffusionDrive~\cite{liao2024diffusiondrive} &C\&L& \underline{98.2} & 96.2 & 82.2 & \textbf{94.7} & \textbf{100} & \underline{88.1} \\
    \midrule
    \textbf{D$^3$-MoE} (Normal) &C  & \textbf{98.3} & \textbf{96.4} & \textbf{82.7} & \underline{94.3} & \underline{99.9} & \textbf{88.2} \\
    \textbf{D$^3$-MoE} (Conservative) &C  & 98.0 & 94.7 & 79.6 & 93.1 & \textbf{100} & 85.5 \\
    \textbf{D$^3$-MoE} (Aggressive) &C  & 96.1 & 91.2 & 79.5 & 87.1 & 99.9 & 80.9 \\
    \midrule
    \textbf{D$^3$-MoE} (Best-of-Three) &C  & 98.7 & 97.7 & 87.6 & 95.3 & 100 & 91.3 \\
    \bottomrule
  \end{tabular}
  \vspace{-0.2cm}
\end{table}

\textbf{Extended PDMS Comparison.} Extended metrics (Tab.~\ref{tab:navsim_v1_1}) further expose these stylistic nuances. The Aggressive style peaks in driving efficiency, validating its proactive longitudinal learning. However, accompanying safety trade-offs reduce its overall EPDMS (76.7) below the Normal (84.3) and Conservative (81.5) baselines. Ultimately, the \textit{Best-of-Three} ensemble yields the highest EPDMS (87.5), confirming adaptive routing as a robust strategy for complex driving scenarios.

\begin{table}[ht]
  \centering
  \vspace{-0.3cm}
  \caption{\textbf{Extended Metrics on Navtest.}}
  \label{tab:navsim_v1_1}
  \setlength{\tabcolsep}{1.5pt}
  \scriptsize
  \resizebox{\columnwidth}{!}{%
  \begin{tabular}{l|ccccccccc>{\columncolor{gray!20}}c}
    \toprule
    \textbf{Model} & \textbf{NC$\uparrow$} & \textbf{DAC$\uparrow$} & \textbf{DDC$\uparrow$} &
    \textbf{TLC$\uparrow$} & \textbf{EP$\uparrow$} & \textbf{TTC$\uparrow$} &
    \textbf{LK$\uparrow$} & \textbf{HC$\uparrow$} & \textbf{EC$\uparrow$} &
    \textbf{EPDMS$\uparrow$} \\
    \midrule
    Ego Status & 93.1 & 77.9 & 92.7 & 99.6 & 86.0 & 91.5 & 89.4 & \textbf{98.3} & 85.4 & 64.0 \\
    TransFuser~\cite{chitta2022transfuser} & 96.9 & 89.9 & 97.8 & \underline{99.7} & 87.1 & 95.4 & 92.7 & \textbf{98.3} & 87.2 & 76.7 \\
    Hydra-MDP++~\cite{li2025hydra++} & 97.2 & \underline{97.5} & \underline{99.4} & 99.6 & 83.1 & 96.5 & 94.4 & \underline{98.2} & 70.9 & 81.4 \\
    DriveSuprim~\cite{yao2026drivesuprim} & 97.5 & 96.5 & \underline{99.4} & 99.6 & \underline{88.4} & 96.6 & 95.5 & \textbf{98.3} & 77.0 & 83.1 \\
    ARTEMIS~\cite{feng2025artemis} & \underline{98.3} & 95.1 & 98.6 & \textbf{99.8} & 81.5 & 97.4 & 96.5 & \textbf{98.3} & \textbf{89.1} & 83.1 \\
    ReCogDrive~\cite{li2025recogdrive} & \underline{98.3} & 95.2 & \textbf{99.5} & \textbf{99.8} & 87.1 & \underline{97.5} & \underline{96.6} & \textbf{98.3} & 86.5 & 83.6 \\
    iPad~\cite{guo2025ipad} & \textbf{98.7} & \textbf{97.8} & 99.1 & \textbf{99.8} & 83.5 & \textbf{98.0} & 96.2 & 98.1 & 85.6 & \underline{84.1} \\
    \midrule
    \textbf{D$^3$-MoE} (Normal) & \underline{98.3} & 96.4 & 98.7 & \textbf{99.8} & 87.8 & 97.3 & \textbf{97.3} & \textbf{98.3} & \underline{88.5} & \textbf{84.3} \\
    \textbf{D$^3$-MoE} (Conservative) & 98.0 & 94.7 & 98.3 & \underline{99.7} & 86.5 & 96.7 & 95.8 & \underline{98.2} & 88.4 & 81.5 \\
    \textbf{D$^3$-MoE} (Aggressive) & 96.1 & 91.2 & 96.8 & 99.5 & \textbf{91.3} & 93.4 & 95.2 & 97.9 & 86.3 & 76.7 \\
    \midrule
    \textbf{D$^3$-MoE} (Best-of-Three) & 98.5 & 97.6 & 99.0 & 99.8 & 91.0 & 97.6 & 98.1 & 98.3 & 91.3 & 87.5 \\
    \bottomrule
  \end{tabular}%
  }
   {\raggedright \scriptsize \\[1.5mm] * All comparison results are based on publicly available data.\par}
 
\end{table}

\textbf{Performance Attribution of the Best-of-Three Ensemble.}
To quantify each style's contribution to the PDMS under the \textit{Best-of-Three} ensemble, we analyzed 12,147 aligned validation samples (Tab.~\ref{tab:style_contribution}). The results reveal strong functional complementarity: despite scoring lower in isolation, the \textit{Aggressive} expert acts as a critical ``ceiling-raiser" by contributing the largest share of optimal solutions (51.52\%). The \textit{Normal} branch provides a stable baseline, securing 45.15\% of the optimums primarily through tied performance, while the \textit{Conservative} branch serves as a safety fallback for highly constrained edge cases (3.33\%). Ultimately, these results confirm that our tri-modal architecture effectively diversifies the solution space to maximize the overall performance ceiling.

\begin{table}[t]
  \centering
  \scriptsize
  \setlength{\tabcolsep}{4pt}
  \caption{Performance Attribution of the Best-of-Three Ensemble.}
  \label{tab:style_contribution}
  \begin{tabular}{l|ccc}
    \toprule
    \textbf{Style Expert} & \textbf{Unique Optimum} & \textbf{Tied Optimum\textsuperscript{*}} & \textbf{Effective Contribution} \\
    \midrule
    Aggressive       & 6,236 & 22    & \textbf{6,258} (\textbf{51.52\%}) \\
    Normal & 1,438 & 4,046 & \textbf{5,484} (\textbf{45.15\%}) \\
    Conservative     & 383 & 22    & \textbf{405} (\textbf{3.33\%}) \\
    \bottomrule
  \end{tabular}
  \vspace{0.1cm}
  
  \raggedright \footnotesize \textsuperscript{*} Ties involving the \textit{Normal} style default to it; the 44 ties occurring solely between \textit{Conservative} and \textit{Aggressive} are split proportionally.
\end{table}

\begin{figure}[t]
  \centering
  \includegraphics[width=1\columnwidth,interpolate=false]{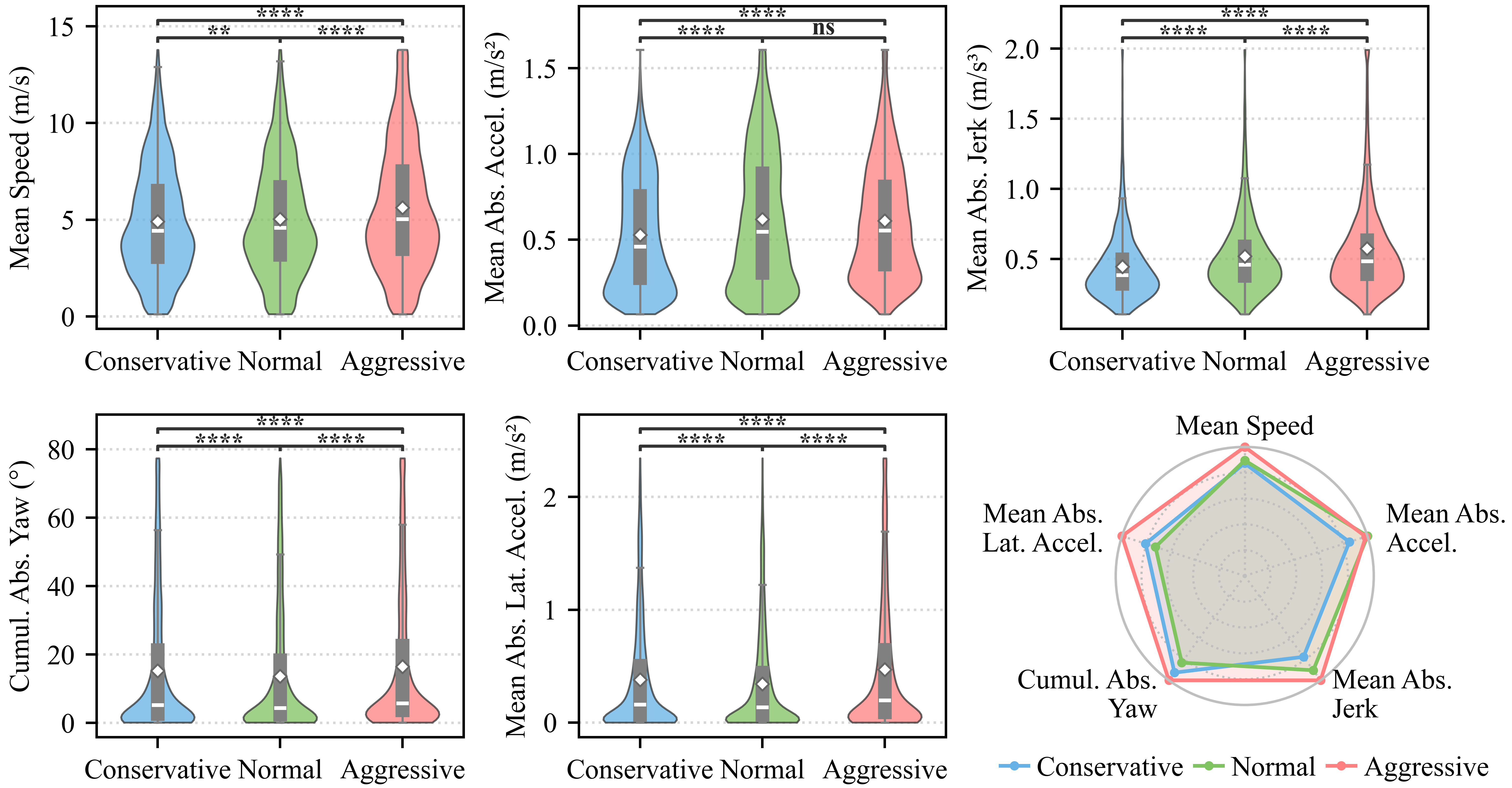}
  \vspace{-0.25cm}
  \caption{Trajectory features across  three styles on the \texttt{navtest} benchmark.}
  \label{fig:dis_inference}
  \vspace{-0.15cm}
\end{figure}

\textbf{Numerical Analysis of Stylistic Kinematics.} 
 To characterize the behavioral differences among the three inference styles on the \texttt{navtest} benchmark, we analyze the statistical distributions of key kinematic features (Fig.~\ref{fig:dis_inference}).
 In terms of longitudinal dynamics, the \textit{Aggressive} style  exhibits the highest means and widest variances in Mean Speed, Mean Abs. Accel., and Mean Abs. Jerk. By forming the outermost boundary on these radar axes, it demonstrates a clear tendency toward proactive acceleration.Conversely, the \textit{Conservative} style suppresses all three metrics in favor of safety.
 Laterally, however, a counter-intuitive but physically grounded pattern emerges: the \textit{Conservative} style records higher Cumul. Abs. Yaw and Mean Abs. Lat. Accel. than the \textit{Normal} baseline. This stems directly from its safety fallback mechanism. While human drivers (\textit{Normal}) naturally smooth paths and cut corners to reduce lateral forces, the \textit{Conservative} strategy rigidly enforces lane-centering. On curved segments, this strict centerline adherence requires frequent steering corrections, inevitably accumulating greater lateral dynamics.

\vspace{-0.4cm}
\subsection{Controllability  Analysis}
\begin{figure*}[!htbp]
  \centering
  \includegraphics[width=0.99\textwidth,interpolate=false]{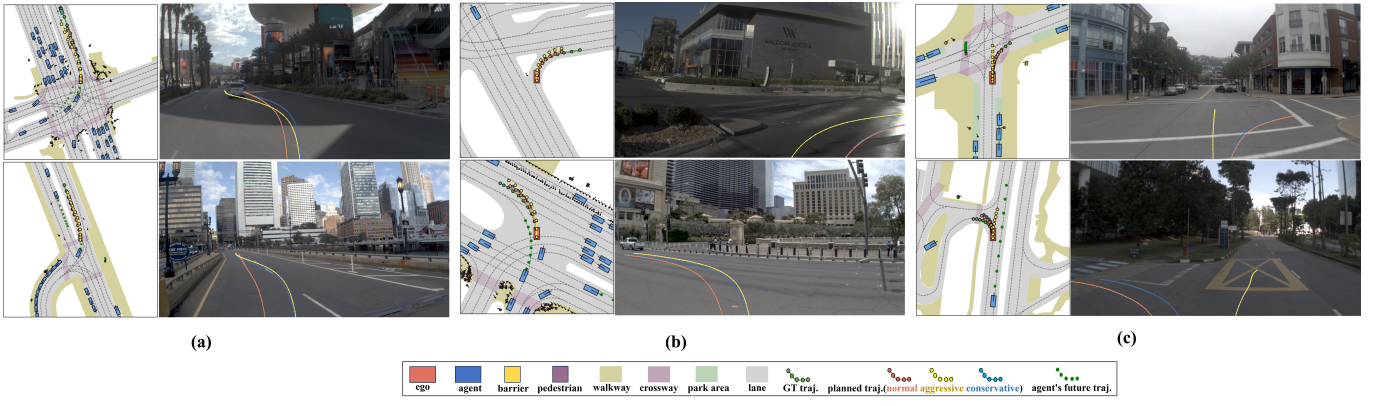}
  \vspace{-0.2cm}
  \caption{Visualization of style-controllable trajectories. Each urban scenario pairs a BEV map (left) with a corresponding front-view image (right). }
  \label{fig:qualitative}
  \vspace{-0.1cm}
\end{figure*}

\textbf{Style-Controllable Trajectory Generation.} 
Fig.~\ref{fig:qualitative} presents qualitative visualizations of our style-conditioned trajectories across diverse urban scenarios, where the three explicit styles exhibit consistent and interpretable motion preferences. The 
\textcolor{aggro_yellow}{\textit{Aggressive}}
 style  favors larger turning radii, higher speeds, and stronger longitudinal progress, occasionally permitting larger temporary lateral offsets before re-centering to maximize passing efficiency. The \textcolor{cons_blue}{\textit{Conservative}} style instead restricts speed and acceleration, yielding cautious, tightly constrained profiles, while the \textcolor{norm_red}{\textit{Normal}} style serves as a balanced baseline between the two extremes.

\begin{figure*}[ht]
  \centering
  \includegraphics[width=0.99\textwidth,interpolate=false]{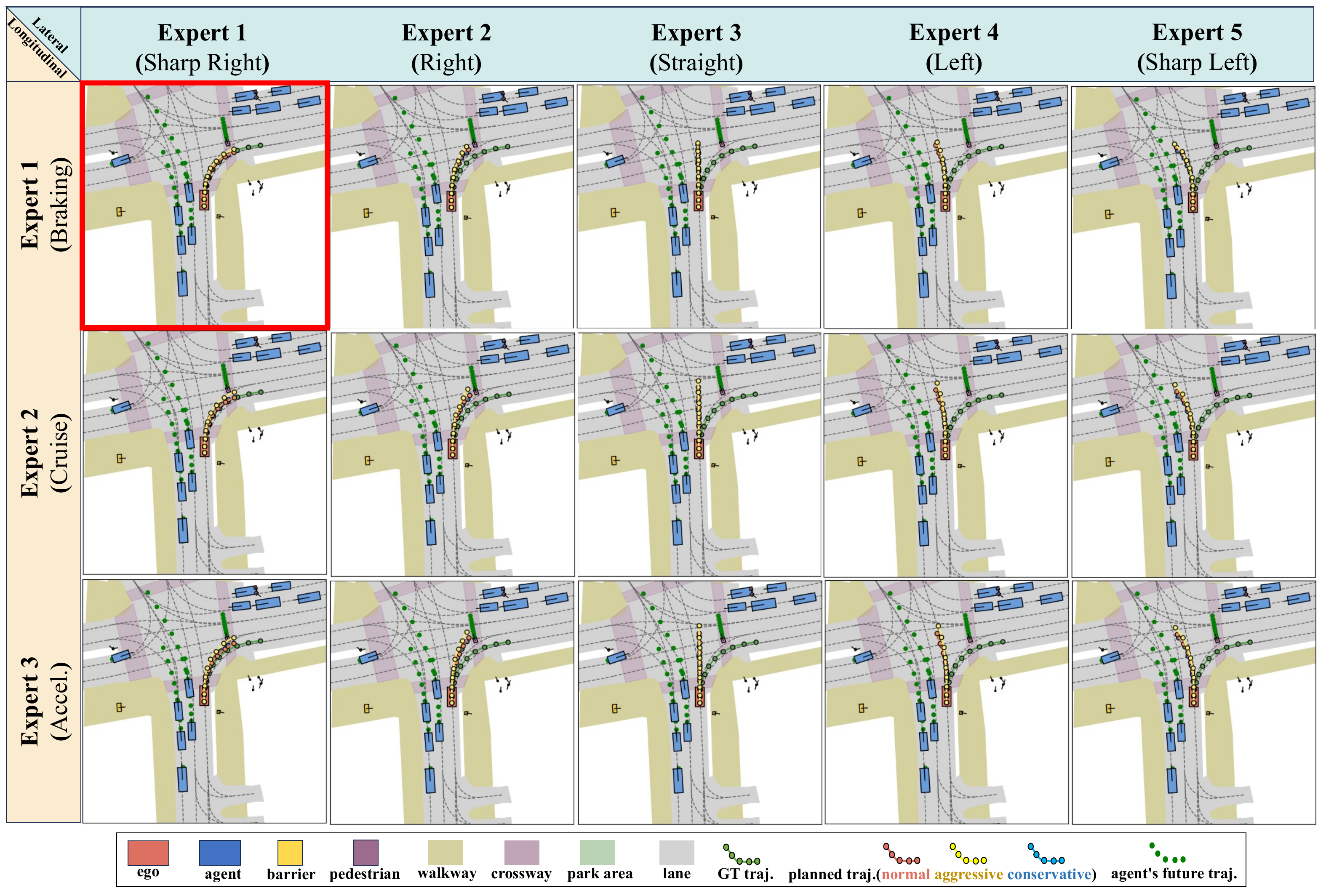}
  \vspace{-0.2cm}
  \caption{Visualization of all longitudinal–lateral expert pairings at an intersection. }
  \label{fig:fixed_expert_pair}
  \vspace{-0.1cm}
\end{figure*}

\textbf{Interpretability of Decoupled Experts.} 
To validate our architecture's physical semantics, we disabled dynamic routing to exhaustively enumerate all $3 \times 5$ combinations of longitudinal (\textit{Braking/Cruise/Accel.}) and lateral (\textit{Sharp Right} to \textit{Sharp Left}) experts (Fig.~\ref{fig:fixed_expert_pair}). The generated tri-modal trajectories reliably align with their assigned semantics, independently dictating speed profiles and steering behaviors. 
Notably, the red bounding box highlights the optimal combination dynamically selected by our router during normal operation (\textit{Braking} + \textit{Sharp Right}), which accurately matches the ground-truth turn, confirming that \textbf{D}$^3$\textbf{-MoE} establishes a highly interpretable, decomposable, and controllable behavioral space.  

\vspace{-0.3cm}
\subsection{Ablation Study}
\textbf{Ablation of Core Components.} 
Tab.~\ref{tab:ablation_module} evaluates our three core designs: \textit{MoE}, \textit{Lateral-Longitudinal Decoupling}, and \textit{Route Supervision}. 
Integrating lateral-longitudinal  decoupling into a monolithic baseline (ID-5) yields a modest gain (ID-4, +0.2 PDMS), mitigating control coupling errors. However, adding dynamic MoE without routing supervision (ID-2) causes a performance collapse (PDMS 86.9), severely degrading passing efficiency (EP 81.4) and safety (TTC 93.8) due to router instability. 

Introducing \textit{Route Supervision} (ID-1) resolves this bottleneck, successfully activating specialized experts to achieve an optimal 88.2 PDMS. Removing decoupling from this full setup (ID-3) drops the PDMS to 87.0 ($-1.2$). 
Together, these results prove that the strength of  \textbf{D}$^3$\textbf{-MoE} lies in the synergy between a decoupled action space and supervised routing.

\begin{table}[t]
  \centering
  \scriptsize
  \setlength{\tabcolsep}{3pt}
  \caption{Ablation of Core Components.}
  \label{tab:ablation_module}
  \resizebox{\columnwidth}{!}{%
  \begin{tabular}{lccc|ccccc>{\columncolor{gray!20}}c}
    \toprule
    \textbf{ID} & \textbf{MoE} & \textbf{Decoupling} & \textbf{Route Supervision} &
    \textbf{NC$\uparrow$} & \textbf{DAC$\uparrow$} & \textbf{EP$\uparrow$} & \textbf{TTC$\uparrow$} & \textbf{C$\uparrow$} & \textbf{PDMS$\uparrow$} \\
    \midrule
    1 & $\checkmark$ & $\checkmark$ & $\checkmark$ &98.3 & 96.4 & 82.7 & 94.3 & 99.9 & 88.2 \\
    2 & $\checkmark$ & $\checkmark$ & $\times$ & 98.1 & 95.3 & 81.4 & 93.8 & 100 & 86.9 \\
    3 & $\checkmark$ & $\times$ & $\checkmark$ & 98.1 & 95.4 & 81.5 & 94.1 & 99.9 & 87.0 \\
    4 & $\times$ & $\checkmark$ & N/A & 98.2 & 95.8 & 82.2 & 94.3 & 99.9 & 87.8 \\
    5 & $\times$ & $\times$ & N/A & 98.0 & 96.2 & 82.4 & 93.7 & 99.9 & 87.6 \\
    \bottomrule
  \end{tabular}
  }
  \vspace{-0.2cm}
\end{table}

\textbf{Ablation of Diffusion Denoising Steps.}
To evaluate the performance-latency trade-off during inference, we ablate the number of diffusion denoising steps $N \in \{1, 2, 3\}$. The recorded inference time denotes the total latency required to generate all three stylistic trajectories in parallel. 

As shown in Tab.~\ref{tab:denoise_steps_tradeoff}, a single denoising step  ($N=1$)  achieves an 88.0 PDMS at minimal latency of 41.14 ms.
 Increasing to $N=2$ peaks the PDMS at 88.2 for a moderate 58.83 ms, while further increasing to $N=3$ strictly incurs computational overhead (69.94 ms) with zero performance gain. 
We therefore adopt $N=2$ to best balances planning quality and inference efficiency.

\begin{table}[t]
  \centering
  \scriptsize
  \setlength{\tabcolsep}{4pt}
  \caption{Ablation on Diffusion Denoising Steps.}
  \label{tab:denoise_steps_tradeoff}
  \resizebox{\columnwidth}{!}{%
  \begin{tabular}{c|ccccc>{\columncolor{gray!20}}c|c}
    \toprule
    \textbf{Denoising Steps} & \textbf{NC$\uparrow$} & \textbf{DAC$\uparrow$} & \textbf{EP$\uparrow$} & \textbf{TTC$\uparrow$} & \textbf{C$\uparrow$} & \textbf{PDMS$\uparrow$} & \textbf{Inference Time (ms)$\downarrow$} \\
    \midrule
    1 & 98.1 & 96.2 & 82.4 & 94.2 & 99.9 & 88.0 & 41.14 \\
    2 & 98.3 & 96.4 & 82.7 & 94.3 & 99.9 & 88.2 & 58.83 \\
    3 & 98.3 & 96.4 & 82.7 & 94.3 & 99.9 & 88.2 & 69.94 \\
    \bottomrule
  \end{tabular}%
  }
  \vspace{-0.2cm}
\end{table}

\textbf{Effect of Dynamic Routing.}
To explicitly verify the effectiveness of our self-supervised routing mechanism, we disabled dynamic selection during inference and exhaustively evaluated all 15 fixed combinations of the lateral ($E_{lat}^1 \dots E_{lat}^5$) and longitudinal ($E_{lon}^1 \dots E_{lon}^3$) experts. As shown in Tab.~\ref{tab:fixed_lat_lon_combo}, relying on any single static pair $(E_{lat}^i, E_{lon}^j)$ triggers a severe performance collapse. Even the best-performing fixed combination ($E_{lat}^4 + E_{lon}^1$) only reaches a 76.09 PDMS, falling drastically short of the full dynamic routing model.
\vspace{-0.2cm}
\begin{table}[ht]
  \centering
  \scriptsize
  \setlength{\tabcolsep} {6pt}
  \caption{PDMS  under Fixed Lateral-Longitudinal Expert Combinations.}
  \label{tab:fixed_lat_lon_combo}
  \begin{tabular}{lccccc}
    \toprule
    \textbf{Longitudinal Expert} & \textbf{Lat-$E_1$} & \textbf{Lat-$E_2$} & \textbf{Lat-$E_3$} & \textbf{Lat-$E_4$} & \textbf{Lat-$E_5$} \\
    \midrule
    \textbf{Lon-$E_1$} & 35.22 & 64.31 & 61.38 & \textbf{76.09} & 38.38 \\
    \textbf{Lon-$E_2$} & 30.95 & 57.06 & 53.34 & 70.48 & 35.27 \\
    \textbf{Lon-$E_3$} & 22.19 & 44.87 & 41.24 & 59.16 & 27.34 \\
    \bottomrule
  \end{tabular}
  \vspace{-0.2cm}
\end{table}

\section{Conclusion}
 In this letter, we presented \textbf{D}$^3$\textbf{-MoE}, a dual-disentangled diffusion Mixture-of-Experts  framework that resolves the style-averaging dilemma in end-to-end autonomous driving. The architecture disentangles trajectory modeling along two complementary axes. Behaviorally, it decouples generation from selection to enable parallel, multi-style candidate synthesis. Physically, it segregates lateral and longitudinal DiT denoisers driven by style-conditioned AdaLN and asymmetric lateral-fusion cross-attention, to seamlessly synergize MoE interpretability with diffusion-based multi-modality. Evaluations on NAVSIM confirm \textbf{D}$^3$\textbf{-MoE} achieves state-of-the-art baseline performance, while our \textit{Best-of-Three} ensemble strategy broadens the solution space to an exceptional 91.3 PDMS. Overall,  \textbf{D}$^3$\textbf{-MoE} provides a robust, interpretable, and controllable paradigm for user-centric multi-modal planning.

\vspace{-0.2cm}
\bibliography{ral2026}

\end{document}